\pdfoutput=1

\documentclass[11pt]{article}

\usepackage[final]{acl}
\usepackage{times}
\usepackage{latexsym}

\usepackage[T1]{fontenc}
\usepackage[utf8]{inputenc}

\usepackage{microtype}

\usepackage{inconsolata}

\usepackage{graphicx}

\usepackage{booktabs}
\usepackage{multirow}
\usepackage{graphicx}
\usepackage{amsmath}
\usepackage{cleveref}
\usepackage{xcolor}

\usepackage{colortbl}
\definecolor{lightblue}{rgb}{0.8,0.85,1}
\definecolor{lightred}{rgb}{1, 0.8, 0.8}
\definecolor{darkgreen}{RGB}{0,150,0}

\usepackage{tcolorbox}
\usepackage{lipsum}  %

\usepackage{CJKutf8}

\title{Fine-Grained and Multi-Dimensional Metrics for Document-Level Machine Translation}

\author{
  Yirong Sun$^{1}$, Dawei Zhu$^{2}$, Yanjun Chen$^{1,3}$,  Erjia Xiao$^{4}$, Xinghao Chen$^{1,3}$, Xiaoyu Shen$^{1}$\thanks{Corresponding authors.}\\
  $^1$ Digital Twin Institute, Eastern Institute of Technology, Ningbo, China\\
  $^2$ Saarland University, Saarland Informatics\\
  $^3$ Department of Computing, The Hong Kong Polytechnic University \\
  $^4$The Hong Kong University of Science and Technology (Guangzhou) \\
  \texttt{win1282467298@gmail.com \qquad xyshen@eitech.edu.cn}
}
\input zhwinfonts
\begin{document}
\maketitle
\begin{abstract}
Large language models (LLMs) have excelled in various NLP tasks, including machine translation (MT), yet most studies focus on sentence-level translation. This work investigates the inherent capability of instruction-tuned LLMs for document-level translation (docMT). Unlike prior approaches that require specialized techniques, we evaluate LLMs by directly prompting them to translate entire documents in a single pass. Our results show that this method improves translation quality compared to translating sentences separately, even without document-level fine-tuning. However, this advantage is not reflected in BLEU scores, which often favor sentence-based translations. We propose using the LLM-as-a-judge paradigm for evaluation, where GPT-4 is used to assess document coherence, accuracy, and fluency in a more nuanced way than n-gram-based metrics. Overall, our work demonstrates that instruction-tuned LLMs can effectively leverage document context for translation. However, we caution against using BLEU scores for evaluating docMT, as they often provide misleading outcomes, failing to capture the quality of document-level translation.\footnote{Our code and the outputs from GPT4-as-a-judge are available at \url{https://github.com/EIT-NLP/BLEUless_DocMT}}

\end{abstract}

\section{Introduction}
Large language models (LLMs) have demonstrated exceptional performance across a wide range of natural language processing tasks~\cite{Radford2019, Brown2020, Touvron2023_llama2, dubey2024llama}. In the realm of machine translation (MT), recent findings also suggest that LLM-based models rival dedicated commercial systems like Google Translate, particularly in translating high-resource languages~\cite{Hendy2023_mt, Peng2023_mt, Jiao2023_gpt_mt,zhu2024fine,Zhu2024_prefmt}. 
Nonetheless, most research has focused only on sentence-level translation. While some studies have begun to explore document-level translation (docMT) with LLMs,  there is a prevailing belief that directly applying instruction-tuned LLMs to docMT performs poorly without specialized training and prompting techniques, largely due to the limited availability of document-level content in instruction-tuning datasets~\cite{wu_doc_level_da,cui2024_doc_mt_context,li2024_doc}. However, their conclusions are frequently drawn from n-gram-based metrics without thorough analysis to substantiate the models' true performance.

In this work, we conduct an in-depth investigation into the \emph{inherent} capabilities of instruction-tuned LLMs in handling docMT tasks. Unlike previous studies that explore special tricks, such as multi-turn inference~\cite{wang-etal-2023-document-level}, we directly prompt LLMs to translate entire documents in a single pass. Comparing this method to a simpler baseline that translates individual sentences separately and then stitches them together, we can evaluate whether instruction-tuned LLMs can leverage their inherent ability to incorporate document-level context and improve translation quality.

A key challenge in our research is the evaluation of document-level machine translation (docMT). Traditional metrics\footnote{While COMET~\cite{rei2020_comet} is more reliable than BLEU for sentence-level translation, it is trained exclusively on sentence-level data. As a result, using COMET to evaluate docMT can be unreliable, since out-of-distribution.} like BLEU\footnote{Although we do not want to use BLEU based metric, it remains a common metric in existing/recent research, despite its limitations.}, ChrF, and TER~\cite{papineni2002bleu,popovic-2015-chrf,snover2006study}, though widely used, often poorly correlate with human judgment~\cite{freitag2022_stop_belu}, especially in docMT, where maintaining coherence and logical flow across a document is essential—something n-gram overlap struggles to capture. Metrics like \textsc{cTT}, \textsc{aZPT}, and \textsc{Blonde}~\cite{jiang2021blonde,wang-etal-2023-document-level} address specific aspects such as terminology consistency and zero-pronoun accuracy, but still rely heavily on word matching and symbolic statistics. We argue that an ideal docMT metric should be (1) context-aware—capturing document-level coherence and accuracy, (2) structured—evaluating aspects such as fluency, accuracy, and coherence separately, and (3) interpretable—explicitly identifying translation errors for clear, objective human evaluation. To this end, we design a GPT-4-based evaluation pipeline to provide deeper insights into the docMT capabilities of LLMs.

\begin{itemize}
  \item We show that translating entire documents yields better results than translating sentences independently then merging them, even without document-level fine-tuning.
  \item We propose using the LLM-as-a-judge paradigm with multiple prompts that assess different aspects of translated text to achieve a more targeted and accurate evaluation.
  \item We recommend against using d-BLEU scores for docMT, as they fail to capture discourse-level phenomena and can often provide misleading results.
\end{itemize}

\section{Problem Settings}
Given a document containing $l$ source sentences $\mathbf{X} = \{x^1, \cdots, x^{l}\}$, the goal of docMT is to generate its translation \(\mathbf{Y} = \{y^1, \cdots, y^{l'}\}\) as a sequence of sentences in the target language. In this work, we explore two approaches for generating translations using instruction-tuned LLMs:
\begin{itemize}
  \item ST[$k$]: We concatenate $k$ source sentences into a chunk, input each chunk into the LLM for translation, and then concatenate the translated chunks together to form the full document translation.
  \item DOC: We instruct the LLM to directly translate the entire document in one pass.
\end{itemize}

The DOC approach is designed to capture inter-sentence dependencies by considering the full document context, potentially leading to more coherent and accurate translations. However, this approach requires the LLM to process and generate longer sequences of text, which can increase the risk of cumulative errors, especially if the model has not been explicitly optimized for document-level translation.

\section{BLEU-based Evaluation}
\label{sec:bleu-based-evaluation}
Document-level BLEU~\citep[d-BLEU,][]{Liu2020_dbleu} is widely used for evaluating translations in DocMT. However, we notice that it is sensitive to overly lengthy generation, which can be problematic as LLMs sometimes overgenerate. We find that even minor overgeneration can significantly affect the final d-BLEU score.\footnote{For completeness, we report results using the standard d-BLEU in Appendix~\ref{appendix:translation_results_d-bleu}.} We argue that documents are generally independent units, so they should be weighted equally in the evaluation. We, therefore, propose an alternative, AvgBLEU, defined as:
\begin{equation*}
\text{AvgBLEU} = \frac{1}{N} \sum_{i=1}^{N} \text{BLEU}\left(Y_{i}^{\text{ref}}, Y_{i}^{\text{pred}}\right)
\end{equation*}
Here, \(N\) is the number of documents, and \(\mathbf{Y}^{\text{ref}}\) and \(\mathbf{Y}^{\text{pred}}\) represent the reference document translations and the predicted translations, respectively. This allows us to calculate the average BLEU score (AvgBLEU) for the entire dataset, providing a comprehensive measure of translation quality.

\begin{table}[h!]
\footnotesize
\centering
\renewcommand{\arraystretch}{0.8}
\scalebox{1}{
\begin{tabular}{@{}lcc@{}}
\toprule[1.0pt]
                  & Number of Sentences & Avg. Document Length \\ \midrule[1.0pt]
            zh-en &   1142              &   252     \\
            en-zh &  1696               &    219     \\
            de-en &   1899              &   204     \\
            en-de &   1780             &    231   \\\midrule[1.0pt]
            Total &    6517             &   225     \\
 \bottomrule[1.0pt]
\end{tabular}
}
\caption{Statistics of our test set. The document length is measured by the token count using Vicuna's tokenizer.}
\label{tab: dataset statistics}
\end{table}

\paragraph{Evaluation Setup.} For evaluation, we use the test set from WMT22~\cite{Kocmi2022_wmt22}, which includes sentence-level reference translations along with annotated document boundaries. Document-level references are obtained by concatenating the corresponding sentence translations. We cover four translation directions in our evaluation: German (\texttt{de}) and Chinese (\texttt{zh}) translated to and from English (\texttt{en}). Specific dataset statistics are presented in Table \ref{tab: dataset statistics}. We evaluate five instruction-tuned LLMs: Vicuna-7B/13B~\cite{Zheng2023_vicuna}, their -16K versions and Mistral-instruct-7B~\cite{Jiang2023_mistral}, all of which have very limited document-level content in their instruction-tuning datasets.

\paragraph{Results.}

Table~\ref{tab:translation_results_d-avgbleu} presents the comparison between the two document-level translation approaches. ST[$k$] consistently achieves higher AvgBLEU scores across all models and nearly all translation directions, with \texttt{zh-en} using Vicuna-7B and Vicuna-13B-16K as the only two exceptions. The specific value of $k$ that yields the highest AvgBLEU score varies depending on the translation direction, however, on average, ST3 achieves the highest score overall. While independently translated sentences yield better AvgBLEU scores than document translations done in one pass by LLMs, manual inspection reveals that ST[$k$] translations often contain more redundancy, literal translations, and disjointed phrasing. While these translations may achieve higher AvgBLEU scores, we find that DOC translations result in more fluent, readable, and cohesive output. This raises concerns about how much AvgBLEU can be trusted as a metric for evaluating docMT.

\begin{table}[t!]
    \scriptsize
    \centering
    \begin{tabular}{lcccccc}
        \toprule[1.1pt]
        \multirow{2}{*}{\textbf{Model}} & \multirow{2}{*}{\textbf{Eval Type}} & \multicolumn{4}{c}{\textbf{Translation Direction}} \\ 
        \cmidrule(l){3-6}
        & & {\textbf{\texttt{zh-en}}} & {\textbf{\texttt{en-zh}}} & {\textbf{\texttt{de-en}}} & {\textbf{\texttt{en-de}}} \\
        \midrule[1.1pt]
        \multirow{4}{*}{Vicuna-7B} & ST1 & 19.70 & 30.97 & 29.42 & 20.82 \\
        & ST2 & 19.69 & 31.65 & \cellcolor{lightred}\bfseries 29.56 & 22.10 \\
        & ST3 & 19.62 & \cellcolor{lightred}\bfseries 32.14 & 29.22 & \cellcolor{lightred}\bfseries 22.53 \\
        & DOC & \cellcolor{lightblue}\bfseries 20.50 & 31.70 & 29.15 & 21.94 \\
        \addlinespace
        \multirow{4}{*}{Vicuna-7B-16K} & ST1 & \cellcolor{lightred}\bfseries 20.26 & 28.08 & 28.16 & 21.11 \\
        & ST2 & 20.05 & 31.17 & 28.78 & \cellcolor{lightred}\bfseries 22.99 \\
        & ST3 & 19.99 & \cellcolor{lightred}\bfseries 31.64 & \cellcolor{lightred}\bfseries 28.89 & 22.93 \\
        & DOC & 20.20 & 30.77 & 28.65 & 21.57 \\
        \addlinespace
        \multirow{4}{*}{Vicuna-13B} & ST1 & \cellcolor{lightred}\bfseries 22.40 & 36.22 & 30.50 & 25.03 \\
        & ST2 & 21.01 & 35.82 & \cellcolor{lightred}\bfseries 30.89 & 25.46 \\
        & ST3 & 21.13 & \cellcolor{lightred}\bfseries 36.24 & 30.84 & \cellcolor{lightred}\bfseries 25.66 \\
        & DOC & 21.83 & 34.93 & 30.60 & 25.59 \\
        \addlinespace
        \multirow{4}{*}{Vicuna-13B-16K} & ST1 &  21.07 & 35.55 & 29.87 & 25.22 \\
        & ST2 & 20.97 & \cellcolor{lightred}\bfseries 36.76 & 30.47 & 24.87 \\
        & ST3 & 20.79 & 36.46 & \cellcolor{lightred}\bfseries 30.71 & \cellcolor{lightred}\bfseries 25.58 \\
        & DOC & \cellcolor{lightblue}\bfseries 21.07 & 34.97 & 30.62 & 25.14 \\
        \addlinespace
        \multirow{4}{*}{Mistral-7B} & ST1 & \cellcolor{lightred}\bfseries 19.82 & 26.24 & 29.23 & 21.28 \\
        & ST2 & 18.89 & 26.84 & \cellcolor{lightred}\bfseries 29.86 & 21.44 \\
        & ST3 & 18.78 & \cellcolor{lightred}\bfseries 26.87 & 29.82 & \cellcolor{lightred}\bfseries 21.74 \\
        & DOC & 18.61 & 24.31 & 28.98 & 21.09 \\
        \bottomrule[1.1pt]
    \end{tabular}
    \caption{AvgBLEU scores with different translation approaches across four translation directions. The best scores are in bold, with red/blue shading indicating the highest score paradigm, respectively. In most cases, merged sentence translations yield higher BLEU scores than direct document translations.}
    \label{tab:translation_results_d-avgbleu}
\end{table}

\section{LLM-as-a-judge Evaluation}
\citet{maruf2021survey} outlines various discourse phenomena that should be considered when evaluating document-level translations, such as cohesion and the use of discourse connectives. In the past, automatic evaluation of these aspects was difficult due to the need for deep semantic understanding, and evaluations typically focused on one aspect at a time using specialized test sets~\cite{hardmeier2010_pronoun, gong2015_cohesion, jwalapuram2019_pronoun}. Inspired by the  ``LLM-as-a-judge” approach~\cite{Zheng2023_vicuna}, we aim to assess multiple aspects simultaneously using a strong LLM. 

\paragraph{Evaluation Setup.} We design four (sub) metrics: (1) \textbf{Fluency}, (2) Content Errors (\textbf{CE}), (3) Lexical Cohesion Errors (\textbf{LE}), and (4) Grammatical Cohesion Errors (\textbf{GE}). All metrics are measured using prompts provided to GPT-4. See Appendix~\ref{appendix:GPT-4 Evaluation Prompt} for details on prompt design.

\textbf{Fluency} is rated on a scale of 1 to 5, with higher being better. Since fluency can be evaluated solely based on the translated text, we present only the model’s outputs to GPT-4 for this assessment, decoupling fluency from metrics that require consideration of source and reference texts.

\textbf{Content Errors} refer to translation mistakes such as mistranslations, omissions, or additions. We instruct GPT-4 (\texttt{gpt-4-0613}) to output a list containing all identified mistakes. The CE score is determined by the length of this list, and report the average CE score over the test set.\footnote{For simplicity, all mistake types are equally weighted, but our approach is flexible and can easily use different weights if certain types are considered more severe than others.} 

\textbf{Cohesion Errors} are further divided into two subcategories: lexical (LE) and grammatical (GE), which affect text connection and the logic flow, respectively. LE includes incorrect vocabulary usage, missing synonyms, or overuse of certain terms that disrupt the flow. GE includes pronouns, conjunctions, and sentence-linking structure mistakes. Similar to CE, we prompt GPT-4 to generate a list of identified errors, with the score corresponding to the length of the list.

Other settings, such as translation directions and the models of interest, remain consistent with Section \ref{sec:bleu-based-evaluation}. Due to the cost associated with using GPT-4, we sample 70 documents per translation direction from the WMT22 dataset for our evaluation.

\begin{figure}[ht!]
    \centering
    \includegraphics[width=1\linewidth]{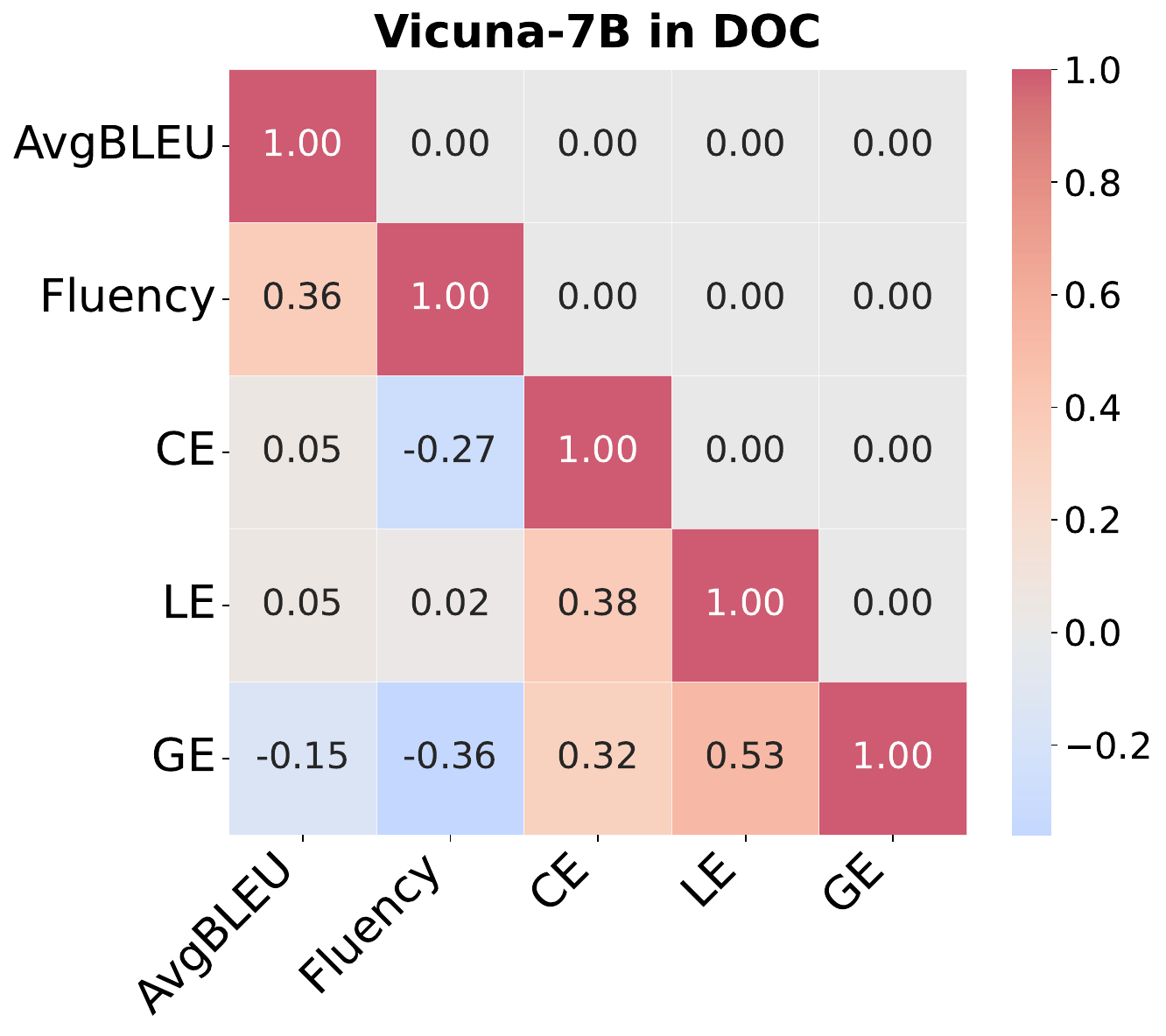}
    \caption{PCC Heatmaps among AvgBLEU, Fluency, CE, LE and GE for Vicuna-7B under DOC evaluation type in the \texttt{en-zh} translation direction.} 
    \label{fig:en-zh_correlation_vicuna7b}
\end{figure}

\paragraph{Results.}
\label{para:LLM-as-a-judge Evaluation-result}
The results with \texttt{en-zh} are shown in Table \ref{tab:gp4-d-avgbleu-en-zh}. Although ST3 scores higher than DOC on AvgBLEU, DOC consistently outperforms ST3 in Fluency. Additionally, DOC generally exhibits fewer CE, also known as content errors. For cohesion errors, the results are mixed: DOC shows better LE with vicuna-7B and its -16K version, and Mistral-7B, while Vicuna-13B and its -16K version yield higher LE. As for GE, DOC performs better with -16K models and Mistral-7B while others are mostly comparable. 
We also observe that the -16K versions perform similarly to their original counterparts in fluency but demonstrate notable improvements in CE reduction. This pattern is consistent across all translation directions, with full results provided in Appendix~\ref{appendix:GPT4-as-a-judge Evaluation Performance}.
Overall, our approach enables a more detailed evaluation of translation quality in DocMT. It clearly shows that instruction-tuned LLMs, even without fine-tuning for document-level MT tasks, are effective at capturing long-context information for DocMT.

To gain a deeper understanding of how these metrics correlates with each other, we compute the Pearson Correlation Coefficients (PCC) among those metrics and visualize them in Figure~\ref{fig:en-zh_correlation_vicuna7b}, as well as translation directions, showing that BLEU score has poor correlation with those discourse-level phenomena metrics. Other translation directions also exhibit low correlation results in Appendix~\ref{appendix:Correlation Visualizations}. Therefore, we suggest not using BLEU score for docMT since it fails to account for discourse-level phenomena, and even worse, it often produces misleading results—such as suggesting that sentence translations are better.

\begin{table}[t!]

\scriptsize
\setlength{\tabcolsep}{3pt}
\centering
\begin{tabular}{lcccccc}
\toprule[1.1pt]
\textbf{Model} & \textbf{Eval Type} & \textbf{AvgBLEU$\uparrow$} & \textbf{Fluency$\uparrow$} & \textbf{CE$\downarrow$} & \textbf{LE$\downarrow$} & \textbf{GE$\downarrow$} \\ 
\midrule[1.1pt]
\multirow{2}{*}{Vicuna-7B}  & ST3 & \cellcolor{lightred}\textbf{33.44} & 3.64 & 4.97 & 2.55 & \cellcolor{lightred}\textbf{1.21} \\
                            & DOC & 28.48 & \cellcolor{lightblue}\textbf{4.04} & \cellcolor{lightblue}\textbf{4.40} & \cellcolor{lightblue}\textbf{2.31} & 1.25 \\
\addlinespace
\multirow{2}{*}{Vicuna-7B-16K} & ST3 & \cellcolor{lightred}\textbf{31.30} & 3.08 & 5.30 & 2.22 & 1.71 \\
                            & DOC & 30.80 & \cellcolor{lightblue}\textbf{3.97} & \cellcolor{lightblue}\textbf{4.72} & \cellcolor{lightblue}\textbf{2.17} & \cellcolor{lightblue}\textbf{1.15} \\
\addlinespace
\multirow{2}{*}{Vicuna-13B} & ST3 & \cellcolor{lightred}\textbf{37.44} & 3.78 & \cellcolor{lightred}\textbf{4.82} & \cellcolor{lightred}\textbf{1.70} & \cellcolor{lightred}\textbf{1.14} \\
                            & DOC & 35.58 & \cellcolor{lightblue}\textbf{4.12} & 4.87 & 2.02 & 1.14 \\
\addlinespace
\multirow{2}{*}{Vicuna-13B-16k} & ST3 & \cellcolor{lightred}\textbf{38.66} & 2.98 & 4.21 & \cellcolor{lightred}\textbf{1.84} & 1.02 \\
                            & DOC & 34.25 & \cellcolor{lightblue}\textbf{4.10} & \cellcolor{lightblue}\textbf{4.15} & 2.04 & \cellcolor{lightblue}\textbf{0.95} \\
\addlinespace
\multirow{2}{*}{Mistral-7B} & ST3 & \cellcolor{lightred}\textbf{26.82} & 2.80 & 6.77 & 4.08 & 2.62 \\
                            & DOC & 23.27 & \cellcolor{lightblue}\textbf{3.11} & \cellcolor{lightblue}\textbf{5.98} & \cellcolor{lightblue}\textbf{3.71} & \cellcolor{lightblue}\textbf{2.51} \\ 
\bottomrule[1.1pt]
\end{tabular}
\caption{Evaluation results (\texttt{en-zh}) by GPT-4 for Vicuna-7B, Vicuna-13B, their -16K versions and Mistral-7B under ST3 and DOC, showing metrics AvgBLEU, fluency, content errors, lexical cohesion errors, and grammatical cohesion errors. Best performances are in bold, with red/blue shading indicating the winning paradigm, respectively.}
\label{tab:gp4-d-avgbleu-en-zh}
\end{table}

\paragraph{Human Agreement.} While some judgments by the LLM-as-a-judge may appear reasonable, certain nuances may still be misinterpreted due to unique human perspectives. To validate the alignment between our LLM-as-a-judge paradigm and human evaluations, we conducted experiments to assess agreement. For each model in both ST3 and DOC, we used 10 samples per translation direction and asked human evaluators to respond with a simple ``yes'' or ``no'' regarding their agreement with the LLM-as-a-judge’s assessments according to our metrics.

Our manual evaluation confirmed a strong alignment between human judgments and the LLM-as-a-judge paradigm. As shown in Table~\ref{tab:agreement_with_human}, GPT-4-as-a-judge achieved approximately 95\% agreement with human evaluations across all languages and evaluation types (ST3 and DOC), indicating robust concordance with human judgment across translation directions and metrics. This high level of agreement further validates GPT-4-as-a-judge as a reliable metric for document-level translation quality.

\begin{table}[h!]
    \centering
    \begin{tabular}{lcccc}
        \toprule[1pt]
         & \textbf{AFluency$\uparrow$} &\textbf{ACE$\uparrow$} & \textbf{ALE$\uparrow$} & \textbf{AGE$\uparrow$} \\
        \midrule[1pt]
        \texttt{zh-en} & 0.96 & 0.95 & 0.94 & 0.96 \\
        \texttt{en-zh} & 0.97 & 0.98 & 0.96 & 0.96 \\
        \texttt{de-en} & 0.98 & 0.96 & 0.94 & 0.95 \\
        \texttt{en-de} & 0.96 & 0.96 & 0.95 & 0.97 \\
        \bottomrule[1pt]
    \end{tabular}
    \caption{Human agreement percentage on GPT4-as-a-judge with our metrics in WMT22. Each judgment is independently reviewed three times by different annotators and consensus results are recorded. AFluency, ACE, ALE, and AGE denote human agreement on the metrics of Fluency, CE, LE, and GE.}
    \label{tab:agreement_with_human}
\end{table}

\begin{figure*}[h!]
\begin{tcolorbox}[colback=gray!10, colframe=black, title=Comparison Cases in ST3 and DOC, center title]
\small

\begin{minipage}{\textwidth}

\begin{center}
    \textbf{Model: Vicuna-7B}
\end{center}
\vspace{0.1em}

\begin{minipage}[b]{0.48\textwidth}
\textbf{Prediction in ST3:}

The 13th Provincial Tea Expo opened today... This morning, the 13th \textcolor{darkgreen}{\textbf{Hunan}} Tea Industry Expo and... were held in the \textcolor{darkgreen}{\textbf{Hunan}} International Convention... 

At the opening ceremony, the provincial leaders awarded the fourth batch of enterprises with the right to use the \textcolor{red}{\textbf{Lake South}} Red Tea trademark. New Hope held the \textcolor{red}{\textbf{Lake}} Red Source. The \textcolor{red}{\textbf{Chaozhou}} tea industry warmly welcomed the arrival of the new spring.
\end{minipage}
\hfill
\vrule
\hfill
\begin{minipage}[b]{0.48\textwidth}
\textbf{Prediction in DOC:}

The 13th \textcolor{darkgreen}{\textbf{Hunan}} Tea Industry Expo opened today at the \textcolor{darkgreen}{\textbf{Hunan}} International Convention...

At the opening ceremony, Hunan's provincial leaders awarded the fourth batch of enterprises with the right to use the \textcolor{darkgreen}{\textbf{Hunan}} Red Tea trademark...
The \textcolor{darkgreen}{\textbf{Hunan}} Red Tea is red in color and has injected the cultural connotation of tea into it, making it popular and lively, and the \textcolor{darkgreen}{\textbf{Hunan}} tea industry is looking forward to a new spring.
\end{minipage}
\end{minipage}

\vspace{0.8em}
\hrule
\vspace{0.8em}

\begin{minipage}{\textwidth}
\begin{center}
    \textbf{Model: Vicuna-7B-16K}
\end{center}
\vspace{0.1em}

\begin{minipage}[t]{0.48\textwidth}
\textbf{Prediction in ST3:}

Color: As shown in the picture (\textcolor{red}{\textbf{please avoid shooting to avoid color difference}}... 

Therefore, girls who can't \textcolor{red}{\textbf{drive}} should not complain about their clothes being old...

\begin{CJK*}{UTF8}{gbsn}
\textcolor{red}{\textbf{2021.6.11部分圈中售出。看好编号下单，古董物品售出不退不换。购买须知}}The products sold at this store are non-refundable...
\end{CJK*}
\textcolor{red}{\textbf{The store does not accept styles that are different from what is imagined}}, and size and style cannot be used as reasons for refunds or exchanges...
\end{minipage}
\hfill
\vrule
\hfill
\begin{minipage}[t]{0.48\textwidth}
\textbf{Prediction in DOC:}

Color: As shown in the picture (\textcolor{darkgreen}{\textbf{Please note that the color difference may not be avoided due to shooting}}...
So, some girls who \textcolor{darkgreen}{\textbf{can't drive vintage clothing}} should not say that the clothes are old-fashioned, but that you are not suitable for it!...

If the item is not suitable for personal reasons, such as not fitting or not liking it, you can ask the store owner to transfer it to the shelf, and \textcolor{darkgreen}{\textbf{once it is sold, it cannot be exchanged or refunded. Part of the circle in the middle was sold on June 11, 2021}}...
\end{minipage}
\end{minipage}
\end{tcolorbox}
\renewcommand{\figurename}{Box} %
\caption{Comparison of Vicuna-7B and Vicuna-7B-16K translations under ST3 and DOC evaluation types in the \texttt{en-zh} translation direction.} 
\label{box:case_study}
\renewcommand{\figurename}{Figure} %
\end{figure*}

\paragraph{Case Study.}
To inspect the advantages of LLMs in docMT, we present two pairs of samples from Vicuna-7B and Vicuna-7B-16K(\texttt{zh-en}), covering beginning, middle, and end of each sample. 

On the right side of first case in Box~\ref{box:case_study}, the translation of ``Hunan” remains consistent throughout the document, illustrating the LLM's capability to leverage context and capture inter-sentence dependencies. Conversely, on the left side, we see an erroneous translation where ``Hunan” is rendered as ``Lake South'' in ``Lake South Red Tea” and simply as ``Lake” in ``Lake Red Source”. Notably, the model in ST3 correctly translates ``Hunan” in other parts of the text. In this case, although ST3 achieves a BLEU score approximately 11.88 points higher than DOC, it is evident that DOC provides more coherent wording and aligns better with natural human expression.

We present another case in Box~\ref{box:case_study}: ST3 translates ``color difference is inevitable in the photos'' as ``please avoid shooting to avoid color difference,” resulting in a significant change in meaning. Additionally, a description about some girls' struggles with a style is mistranslated as ``girls who can’t drive” where ``drive” is incorrectly used as an intransitive verb. In contrast, DOC accurately translates this as ``some girls who can’t drive vintage clothing” preserving the intended meaning while employing the same words in different contexts. Furthermore, the statement ``once it is sold, it cannot be exchanged or refunded. Part of the circle in the middle was sold on June 11, 2021” is correctly translated in DOC, while ST3 reject translating this segment entirely. These cases explicitly demonstrate that instruction-tuned LLMs can effectively capture inter-sentence dependencies by considering the entire document context, leading to a deeper understanding of the text and fewer content errors.

Thus, we advocate against using BLEU as an evaluation metric for docMT, as it fails to detect the true advantages of LLMs in this context and can yield misleading results.

\section{Conclusion}
In this work, we investigate the performance of instruction-tuned LLMs in document-level machine translation (docMT), comparing the translation of entire documents in a single pass to translating individual sentences that are then concatenated. Our findings show that translating entire documents yields better results, as the model can capture inter-sentence dependencies and maintain discourse coherence, even without explicit fine-tuning for docMT tasks.
However, evaluating these improvements is challenging. Traditional metrics like d-BLEU fail to consider discourse-level phenomena, often favoring sentence-level translations and producing misleading results. To address this limitation, we propose the LLM-as-a-judge approach, utilizing GPT-4 to assess specific aspects of discourse through tailored prompts. 
This method enhances interpretability and can be adapted for evaluating translation quality in other domains.

\section*{Limitations}

\paragraph{Translation Directions.}
We evaluate only high-resource language pairs, which limits the generalizability of our findings for low-resource languages. Due to data availability constraints, our experiments focus on well-resourced translation directions. Future research should explore whether instruction-tuned LLMs translating entire documents yield better results than translating sentences independently in low-resource languages.

\paragraph{Model Size and Diversity.} We focus exclusively on small-scale LLMs. Future work should investigate larger models to observe whether instruction-tuned LLMs continue to perform better in docMT, and whether BLEU would work.

\paragraph{Max Length.}
A small fraction ($\sim 2\%$) of documents in WMT22, including both their source texts and translations, exceed 2048 tokens. Thus, we focus solely on samples within the model’s context length (2048 tokens), as these instruction-tuned LLMs are primarily trained on text within this limit. In future work, we will evaluate LLMs with longer context lengths, examine -16K models, and investigate  whether long conversation instruction-tuned will help and whether those phenomena persist when translating text that exceeds the models' context length.

\section*{Ethical Considerations}  
Our study aims to investigate the docMT reliability of instruction-tuned LLMs without fine-tuning for docMT, concerned by the potential for accumulating errors during decoding, which may lead to increased hallucinations. We expect minimal social risks associated with our efforts. 

\bibliography{custom}

\appendix

\section{Evaluation Metrics Shortcoming Analysis}
While COMET has been shown to provide more reliable evaluations than BLEU in many cases, it is primarily trained on sentence-level translations and, as such, is not well-suited for docMT. Given that COMET lacks specific training to capture the complexities of inter-sentence dependencies and discourse-level phenomena, it is not an ideal metric for evaluating the true capabilities of LLMs in docMT tasks. Therefore, in this work, we opted to explore more appropriate evaluation methods tailored to document-level translation challenges.

Similarly, metrics like ChrF~\cite{popovic-2015-chrf}, ChrF2, and TER have made incremental progress by incorporating word-level matching mechanisms that extend beyond simple token overlap, but they still fundamentally rely on surface-level statistics. Like BLEU, these metrics do not adequately account for deeper discourse relationships, cohesion, and the broader context required for accurate docMT assessment. As a result, their limitations become more apparent when evaluating LLMs on longer texts, where capturing the overall document structure is essential.

While metrics such as \textsc{cTT} and \textsc{aZPT} are designed to address specific issues like terminology consistency and zero pronoun accuracy, they remain grounded in automatic identified lexical alignment. These metrics operate under the assumption that the presence of specific terminology or pronouns directly correlates with translation quality. However, in practice, meaning can be conveyed in multiple ways without strictly adhering to these surface-level features. This makes \textsc{cTT} and \textsc{aZPT} limited in scope, as they are unable to fully assess translation quality when alternative phrasing or omitted pronouns still preserve meaning accurately.

Blonde represents a more sophisticated approach by categorizing and analyzing discourse coherence using linguistic features such as verb tense (e.g., \texttt{VBD} for past tense verbs). While this is a step toward capturing discourse-level phenomena, Blonde is still constrained by symbolic statistical methods. Its reliance on predefined linguistic categories means that it struggles to account for the full range of discourse phenomena that can arise in real-world documents. As a result, these metrics, despite their improvements, remain insufficient for capturing the nuances of document-level translation in its entirety.

To address these limitations, we propose leveraging LLM-as-a-judge for evaluating docMT. By employing GPT-4 with specifically designed judging prompts, we can define and assess discourse phenomena in a more abstract and flexible manner, similar to how human evaluators would approach the task. This method avoids the need to predefine all possible linguistic cases and allows for a more holistic evaluation of translation quality, ensuring that complex discourse relationships and contextual dependencies are properly recognized. In doing so, we provide more reliable and interpretable metrics and prompts for evaluating document-level translations, moving beyond the restrictive frameworks of traditional metrics.

\section{d-BLEU Performance}
\label{appendix:translation_results_d-bleu}
We observe that the trend in Table~\ref{tab:d-bleu_result_in_appendix} remains consistent with Table~\ref{tab:translation_results_d-avgbleu}, and the BLEU score shows an even stronger preference for translations that are processed separately and concatenated. 
It is worth to Notice that the red data point in Table~\ref{tab:d-bleu_result_in_appendix} is influenced by the sensitivity of BLEU, where a certain generated translation contains a long-repeated incorrect token toward the end, thus lowering the overall score. When calculating the BLEU score for this sample, we find that the document receives a score near zero, despite the fact that the earlier part of the translation is mostly accurate. This sensitivity is one of the reasons why BLEU should not be used in docMT.
 
\begin{table*}[h!]
    \centering
    \setlength{\tabcolsep}{7mm}{
    \begin{tabular}{lccccc}
        \toprule[1.1pt]
        \multirow{2}{*}{\textbf{Model}} & \multirow{2}{*}{\textbf{Eval Type}} & \multicolumn{4}{c}{\textbf{Translation Direction}} \\ 
        \cmidrule(l){3-6}
        & & \textbf{\texttt{zh-en}} & \textbf{\texttt{en-zh}} & \textbf{\texttt{de-en}} & \textbf{\texttt{en-de}} \\
        \midrule[1.1pt]
        \multirow{4}{*}{Vicuna-7B} 
        & ST1 & 18.75 & 32.43 & 30.00 & 21.96 \\
        & ST2 & 19.99 & 33.52 & \cellcolor{lightred}\textbf{30.87} & 23.35 \\
        & ST3 & \cellcolor{lightred}\textbf{20.52} & \cellcolor{lightred}\textbf{33.92} & 30.68 & \cellcolor{lightred}\textbf{23.96} \\
        & DOC & 19.93 & 32.40 & 30.27 & 22.90 \\
        \addlinespace
        \multirow{4}{*}{Vicuna-7B-16K} 
        & ST1 & 19.54 & 28.45 & 29.75 & 21.49 \\
        & ST2 & 20.38 & 32.52 & \cellcolor{lightred}\textbf{30.60} & \cellcolor{lightred}\textbf{24.27} \\
        & ST3 & \cellcolor{lightred}\textbf{20.43} & \cellcolor{lightred}\textbf{33.15} & 30.56 & 23.95 \\
        & DOC & 19.50 & 30.87 & \textcolor{red}{16.58} & 21.34 \\
        \addlinespace
        \multirow{4}{*}{Vicuna-13B} 
        & ST1 & 21.33 & 37.62 & 31.98 & 26.24 \\
        & ST2 & 21.26 & 37.70 & \cellcolor{lightred}\textbf{32.16} & \cellcolor{lightred}\textbf{27.19} \\
        & ST3 & \cellcolor{lightred}\textbf{21.89} & \cellcolor{lightred}\textbf{37.97} & 32.15 & 26.94 \\
        & DOC & 21.63 & 35.87 & 31.22 & 26.31 \\
        \addlinespace
        \multirow{4}{*}{Vicuna-13B-16K} 
        & ST1 & 21.22 & 37.29 & 31.48 & 26.21 \\
        & ST2 & 21.99 & \cellcolor{lightred}\textbf{37.93} & 31.78 & 26.53 \\
        & ST3 & \cellcolor{lightred}\textbf{22.61} & 37.83 & \cellcolor{lightred}\textbf{32.02} & \cellcolor{lightred}\textbf{26.98} \\
        & DOC & 21.84 & 35.01 & 31.60 & 26.03 \\
        \addlinespace
        \multirow{4}{*}{Mistral-7B} 
        & ST1 & 18.69 & 25.75 & 29.50 & 22.37 \\
        & ST2 & \cellcolor{lightred}\textbf{19.29} & \cellcolor{lightred}\textbf{26.83} & 30.02 & 21.98 \\
        & ST3 & 18.82 & 26.81 & \cellcolor{lightred}\textbf{30.11} & \cellcolor{lightred}\textbf{22.60} \\
        & DOC & 13.70 & 17.54 & 27.50 & 21.98 \\
        \bottomrule[1.1pt]
    \end{tabular}}
    \caption{d-BLEU score with different translation paradigms. More explanations about not using d-BLEU and about the red data point in the Table are stated in~\Cref{appendix:translation_results_d-bleu}}
    \label{tab:d-bleu_result_in_appendix}
\end{table*}

\section{GPT4-as-a-judge Evaluation Prompts}
\label{appendix:GPT-4 Evaluation Prompt}
\subsection{Fluency}
Fluency refers to the naturalness and smoothness of a text in the target language, without awkward or unnatural phrasing. In machine translation evaluation, fluency is crucial for assessing the readability and linguistic quality of the output, which is often not fully captured by traditional metrics like BLEU. While BLEU focuses on n-gram overlap between the translation and reference text, it does not directly evaluate how natural the translation sounds or whether it adheres to syntactic rules. Fluency, in contrast, provides a more nuanced evaluation of the model's ability to produce human-like text.

In this task, we assess fluency on a scale of 1 to 5, with higher scores indicating more fluent translations. Evaluators are instructed to analyze the text and assign a score based solely on the naturalness and grammatical correctness of the model's output. 

Importantly, the fluency evaluation is conducted in isolation, decoupled from cohesion, with only inference text input, to ensure a clear focus on the text’s immediate readability. Cohesion, which refers to the grammatical and lexical connectivity between text units~\cite{halliday2014cohesion}, is considered separately to avoid confounding the two metrics, as fluency and cohesion could be correlated, as it is common sense that if a text is cohesive, its flow is naturally better. See the correlation heatmaps like Figure~\ref{fig:en-zh_correlation_vicuna7b} which show that our prompt design successfully decouples these two metrics.

The evaluation is supported by specific examples and justifications for the assigned score. Below is the prompt used to guide the evaluation:

\begin{quote}
\textit{Please evaluate the fluency of the following text in the target language (English, Chinese, or German).}

\textbf{Instructions:}
\begin{itemize}
    \item \textbf{Task}: Evaluate the fluency of the text.
    \item \textbf{Scoring}: Provide a score from 1 to 5, where:
    \begin{itemize}
        \item \textbf{5}: The text is highly fluent, with no grammatical errors, unnatural wording, or stiff syntax.
        \item \textbf{4}: The text is mostly fluent, with minor errors that do not impede understanding.
        \item \textbf{3}: The text is moderately fluent, with noticeable errors that may slightly affect comprehension.
        \item \textbf{2}: The text has low fluency, with frequent errors that hinder understanding.
        \item \textbf{1}: The text is not fluent, with severe errors that make it difficult to understand.
    \end{itemize}
    \item \textbf{Explanation}: Support your score with specific examples to justify your evaluation.
\end{itemize}

\textbf{Output Format:}
\begin{quote}
Provide your evaluation in the following JSON format:

\{ "Fluency": \{ "Score": "\textless the score\textgreater", "Explanation": "\textless your explanation on how you made the decision\textgreater" \} \}
\end{quote}

\textbf{Text to Evaluate:}

\textit{"{inference text}"}
\end{quote}

\subsection{Content Errors}
Unlike fluency, which assesses the naturalness and grammatical correctness of the output, accuracy focuses on the semantic alignment between the translated text and the original reference. The evaluator’s task is to identify and categorize errors that affect the translation’s fidelity, such as mistranslations, omissions, or additions.

Rather than relying on simple n-gram matching, the evaluation emphasizes meaning preservation. The evaluator compares the translation with the reference text, identifying instances where the translation deviates in meaning. However, if the translated text conveys the same information as the reference but uses different words or phrasing, it is not considered an error, since we suspect that this phenomenon could happen in LLMs in document-level translation task. This approach ensures that the model's output is evaluated based on its ability to faithfully represent the source content, capturing specific issues like mistranslations or information loss, and ensuring semantic integrity.
The accuracy evaluation prompt is structured as follows:

\begin{quote}
\textit{Please evaluate the accuracy of the following text by comparing it to the reference text provided.}

\textbf{Instructions:}
\begin{itemize}
    \item \textbf{Task}: Compare the text to the reference text.
    \item \textbf{Identify Mistakes}: List all mistakes related to accuracy.
    \begin{itemize}
        \item \textbf{Mistake Types}:
        \begin{itemize}
            \item \textbf{Wrong Translation}: Incorrect meaning or misinterpretation leading to wrong information.
            \item \textbf{Omission}: Missing words, phrases, or information present in the reference text.
            \item \textbf{Addition}: Extra words, phrases, or information not present in the reference text.
            \item \textbf{Others}: Mistakes that are hard to define or categorize.
        \end{itemize}
    \end{itemize}
    \item \textbf{Note}: If the text expresses the same information as the reference text but uses different words or phrasing, it is \textbf{not} considered a mistake.
    \item \textbf{Provide a List}: Summarize all mistakes without repeating the exact sentences. Provide an empty list if there are no mistakes.
\end{itemize}
\end{quote}

\textbf{Output Format:}
\begin{quote}
\{
  "Accuracy": \{
    "Mistakes": [
      "<list of all mistakes in the text, provide an empty list if there are no mistakes>"
    ]
  \}
\end{quote}

\textbf{Reference Text:}

\textit{"{reference text}"}

\textbf{Text to Evaluate:}

\textit{"{inference text}"}

\subsection{Cohesion Errors}
Cohesion is a critical aspect of machine translation evaluation as it ensures that the various parts of the text are well-connected and that the overall flow is logical. Unlike metrics such as fluency or accuracy, cohesion specifically examines how sentences are linked together through lexical (lexical cohesion) and grammatical (grammatical cohesion) means~\cite{maruf2021survey}. This is particularly important in document-level translation, where the consistency of vocabulary and the logical connection of grammatical structures across a longer text are challenging for models to maintain.

In the context of translations produced using the ST3 and DOC paradigms, evaluating cohesion allows us to assess whether the model effectively leverages contextual information to maintain consistency across the text. By decoupling cohesion from fluency, our evaluation framework enables evaluators to focus specifically on identifying lexical cohesion mistakes—such as incorrect vocabulary usage, missing synonyms, or overuse of certain terms that disrupt the flow—and grammatical cohesion mistakes—such as errors in pronouns, conjunctions, or sentence-linking structures.

The evaluator is asked to identify any mistakes related to cohesion and categorize them as either lexical or grammatical cohesion issues. The evaluation prompt is structured as follows:

\begin{quote}
\textit{Please evaluate the cohesion of the following text by comparing it to the reference text.}

\textbf{Instructions:}
\begin{itemize}
    \item \textbf{Task}: Evaluate the cohesion of the text.
    \item \textbf{Definition}: Cohesion refers to how different parts of a text are connected using language structures like grammar and vocabulary. It ensures that sentences flow smoothly and the text makes sense as a whole.
    \item \textbf{Identify Mistakes}: List all mistakes related to cohesion.
    \begin{itemize}
        \item \textbf{Lexical Cohesion Mistakes}: Issues with vocabulary usage, incorrect or missing synonyms, or overuse of certain words that disrupt the flow.
        \item \textbf{Grammatical Cohesion Mistakes}: Problems with pronouns, conjunctions, or grammatical structures that link sentences and clauses.
    \end{itemize}
    \item \textbf{Provide Lists}: Provide separate lists for lexical cohesion mistakes and grammatical cohesion mistakes. Provide empty lists if there are no mistakes.
\end{itemize}
\end{quote}

\textbf{Output Format:}
\begin{quote}
\{
  "Cohesion": \{
    "Lexical Cohesion Mistakes": [
      "\textless list of all mistakes in the text, provide an empty list if there are no mistakes\textgreater"
    ],
    "Grammatical Cohesion Mistakes": [
      "\textless list of all mistakes in the text, provide an empty list if there are no mistakes\textgreater"
    ]
  \}\}
\end{quote}

\textbf{Reference Text:}

\textit{"{reference text}"}

\textbf{Text to Evaluate:}

\textit{"{inference text}"}

\newpage
\onecolumn
\section{GPT4-as-a-judge Evaluation Performance}
\label{appendix:GPT4-as-a-judge Evaluation Performance}
\begin{table*}[h!]
\centering
\setlength{\tabcolsep}{4.75mm}{
\begin{tabular}{lcccccc}
\toprule[1.1pt]
\textbf{Model} & \textbf{Eval Type} & \textbf{AvgBLEU$\uparrow$} & \textbf{Fluency$\uparrow$} & \textbf{CE$\downarrow$} & \textbf{LE$\downarrow$} & \textbf{GE$\downarrow$} \\ 
\midrule[1.1pt]
\multirow{2}{*}{Vicuna-7B} 
 & ST3 & \cellcolor{lightred}\textbf{20.25} &  \cellcolor{lightred}\textbf{4.27}  &  \cellcolor{lightred}\textbf{3.06}  & 1.46 & 0.96 \\
 & DOC & 19.36 & 4.20 & 3.24 & \cellcolor{lightblue}\textbf{1.40} & \cellcolor{lightblue}\textbf{0.67} \\ 
\addlinespace
\multirow{2}{*}{Vicuna-7B-16K} 
 & ST3 &  20.07  &  4.24  &   3.50   & 1.20 & 0.77 \\
 & DOC & \cellcolor{lightblue}\textbf{21.14} & \cellcolor{lightblue}\textbf{4.38} & \cellcolor{lightblue}\textbf{3.24} & \cellcolor{lightblue}\textbf{1.18} & \cellcolor{lightblue}\textbf{0.67} \\ 
\addlinespace
\multirow{2}{*}{Vicuna-13B}
 & ST3 & 22.46 & 4.09 & 3.31 & 1.57 & 1.07 \\
 & DOC &   \cellcolor{lightblue}\textbf{23.46}  & \cellcolor{lightblue}\textbf{4.34} & \cellcolor{lightblue}\textbf{3.04} & \cellcolor{lightblue}\textbf{1.04} & \cellcolor{lightblue}\textbf{0.59} \\ 
\addlinespace
\multirow{2}{*}{Vicuna-13B-16K} 
 & ST3 &  21.80  & 4.21   &  3.27   & 0.90 & 0.65 \\
 & DOC & \cellcolor{lightblue}\textbf{22.22} & \cellcolor{lightblue}\textbf{4.48} & \cellcolor{lightblue}\textbf{2.82} & \cellcolor{lightblue}\textbf{0.80} & \cellcolor{lightblue}\textbf{0.41} \\ 
\addlinespace
\multirow{2}{*}{Mistral-7B}
 & ST3 & 18.84 & 3.96 & 4.41 & 1.47 & 1.10 \\
 & DOC &  \cellcolor{lightblue}\textbf{19.50}  & \cellcolor{lightblue}\textbf{4.34} & \cellcolor{lightblue}\textbf{3.50} & \cellcolor{lightblue}\textbf{1.30} & \cellcolor{lightblue}\textbf{0.83} \\ 
\bottomrule[1.1pt]
\end{tabular}}
\caption{Evaluation results (\texttt{zh-en}) by GPT-4 for Vicuna-7B, Vicuna-13B, and Mistral-7B under ST3 and DOC evaluation types, showing metrics AvgBLEU, Fluency, Content Errors(CE), Lexical Cohesion Errors(LE), and Grammatical Cohesion Errors(GE).}
\end{table*}

\begin{table*}[h!]
\setlength{\tabcolsep}{4.75mm}
\centering
\begin{tabular}{lcccccc}
\toprule[1.1pt]
\textbf{Model} & \textbf{Eval Type} & \textbf{AvgBLEU$\uparrow$} & \textbf{Fluency$\uparrow$} & \textbf{CE$\downarrow$} & \textbf{LE$\downarrow$} & \textbf{GE$\downarrow$} \\ 
\midrule[1.1pt]
\multirow{2}{*}{Vicuna-7B}  & ST3 & \cellcolor{lightred}\textbf{24.53} & 3.23 & 7.31 & \cellcolor{lightred}\textbf{3.73} & 3.00 \\
                            & DOC & 21.18 & \cellcolor{lightblue}\textbf{3.61} & \cellcolor{lightblue}\textbf{6.46} & 3.76 & \cellcolor{lightblue}\textbf{2.87} \\
\addlinespace
\multirow{2}{*}{Vicuna-7B-16K}  & ST3 & 26.02 & \cellcolor{lightred}\textbf{4.21} & 2.88 & 1.17 & 0.77 \\
                            & DOC & \cellcolor{lightblue}\textbf{26.95} &  4.11 & \cellcolor{lightblue}\textbf{2.84} & \cellcolor{lightblue}\textbf{0.98} & \cellcolor{lightblue}\textbf{0.67} \\
\addlinespace
\multirow{2}{*}{Vicuna-13B} & ST3 & 26.76 & 3.59 & 6.84 & 3.79 & 2.39 \\
                            & DOC &  \cellcolor{lightblue}\textbf{27.32}  & \cellcolor{lightblue}\textbf{3.90} & \cellcolor{lightblue}\textbf{5.23} & \cellcolor{lightblue}\textbf{3.34} & \cellcolor{lightblue}\textbf{1.96} \\
\addlinespace
\multirow{2}{*}{Vicuna-13B-16K} & ST3 & 28.15 & 4.32 & 2.67 & \cellcolor{lightred}\textbf{0.78} & \cellcolor{lightred}\textbf{0.42} \\
                            & DOC &  \cellcolor{lightblue}\textbf{28.54}  & \cellcolor{lightblue}\textbf{4.45} & \cellcolor{lightblue}\textbf{2.28} &  0.95 & 0.45  \\
\addlinespace
\multirow{2}{*}{Mistral-7B} & ST3 & \cellcolor{lightred}\textbf{22.32} & 3.16 & 6.83 & 4.64 &  \cellcolor{lightred}\textbf{2.93}  \\
                            & DOC & 21.46 & \cellcolor{lightblue}\textbf{3.17} & \cellcolor{lightblue}\textbf{6.63} & \cellcolor{lightblue}\textbf{4.51} & 2.96 \\ 
\bottomrule[1.1pt]
\end{tabular}
\caption{Evaluation results (\texttt{en-de}) by GPT-4 for Vicuna-7B, Vicuna-13B, and Mistral-7B under ST3 and DOC evaluation types, showing metrics AvgBLEU, Fluency, Content Errors(CE), Lexical Cohesion Errors(LE), and Grammatical Cohesion Errors(GE).}
\end{table*}

\begin{table*}[h!]
\centering
\setlength{\tabcolsep}{4.75mm}{
\begin{tabular}{lcccccc}
\toprule[1.1pt]
\textbf{Model} & \textbf{Eval Type} & \textbf{AvgBLEU$\uparrow$} & \textbf{Fluency$\uparrow$} & \textbf{CE$\downarrow$} & \textbf{LE$\downarrow$} & \textbf{GE$\downarrow$} \\ 
\midrule[1.1pt]
\multirow{2}{*}{Vicuna-7B}  & ST3 & \cellcolor{lightred}\textbf{26.82} & 4.11 & 4.01 &  \cellcolor{lightred}\textbf{1.23}  & 1.11 \\
                            & DOC & 25.64 & \cellcolor{lightblue}\textbf{4.31} & \cellcolor{lightblue}\textbf{3.14} & 1.67 & \cellcolor{lightblue}\textbf{0.66} \\
\addlinespace
\multirow{2}{*}{Vicuna-7B-16k}  & ST3 & \cellcolor{lightred}\textbf{23.56} & \cellcolor{lightred}\textbf{3.61} & \cellcolor{lightred}\textbf{5.74} &   3.52  & 2.52 \\
                            & DOC & 21.71 & 3.54  &  5.81  & \cellcolor{lightblue}\textbf{3.47} & \cellcolor{lightblue}\textbf{2.21} \\
\addlinespace
\multirow{2}{*}{Vicuna-13B} & ST3 & 27.23 & 4.30 &  \cellcolor{lightred}\textbf{3.06}  &  \cellcolor{lightred}\textbf{1.13}  & 0.66 \\
                            & DOC &  \cellcolor{lightblue}\textbf{28.44}  & \cellcolor{lightblue}\textbf{4.33} & 3.36 & 1.33 & \cellcolor{lightblue}\textbf{0.60} \\
\addlinespace
\multirow{2}{*}{Vicuna-13B-16K}  & ST3 & \cellcolor{lightred}\textbf{26.55} & 4.15 & 5.47 &  \cellcolor{lightred}\textbf{2.72}  & \cellcolor{lightred}\textbf{1.91} \\
                            & DOC & 26.28 & \cellcolor{lightblue}\textbf{4.18} & \cellcolor{lightblue}\textbf{4.7} & 2.91 & 1.92 \\
\addlinespace
\multirow{2}{*}{Mistral-7B} & ST3 & \cellcolor{lightred}\textbf{26.09} & 4.10 &  \cellcolor{lightred}\textbf{4.73}  & 1.49 & 1.37 \\
                            & DOC & 25.68 & \cellcolor{lightblue}\textbf{4.33} & 4.89 & \cellcolor{lightblue}\textbf{1.26} & \cellcolor{lightblue}\textbf{0.80} \\ 
\bottomrule[1.1pt]
\end{tabular}}
\caption{Evaluation results (\texttt{de-en}) by GPT-4 for Vicuna-7B, Vicuna-13B, and Mistral-7B under ST3 and DOC evaluation types, showing metrics AvgBLEU, Fluency, Content Errors(CE), Lexical Cohesion Errors(LE), and Grammatical Cohesion Errors(GE).}
\end{table*}

\newpage
\section{Correlation Visualizations}
\label{appendix:Correlation Visualizations}

\begin{figure*}[ht!]
    \centering
    \includegraphics[width=1\linewidth]{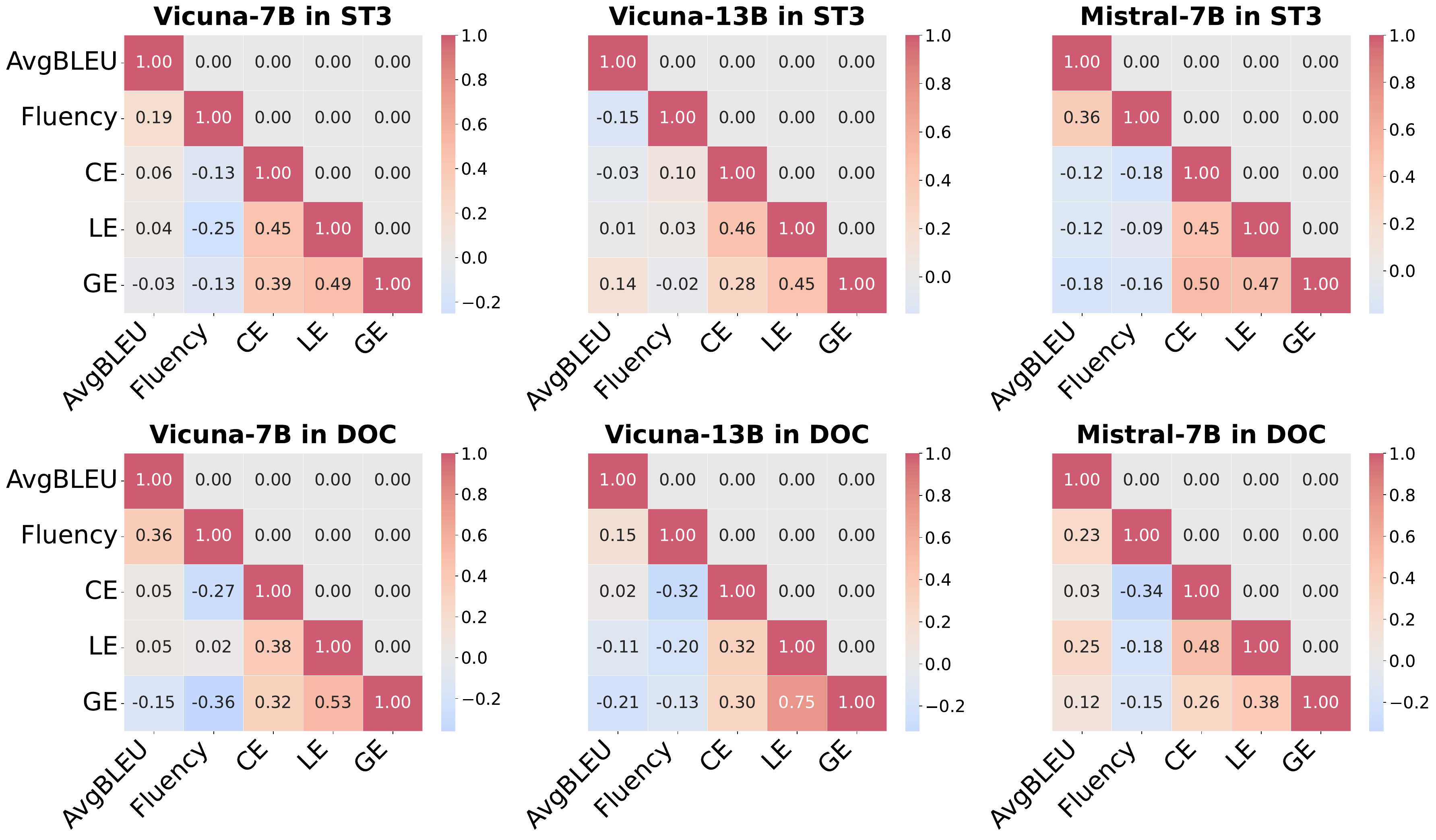}
    \caption{PCC Heatmaps among AvgBLEU, Fluency, CE, LE, GE for Vicuna-7B, Vicuna-13B, and Mistral-7B under ST3 and DOC evaluation types in translation direction of \texttt{en-zh}.} 
\end{figure*}

\begin{figure*}[h!]
    \centering
    \includegraphics[width=1\linewidth]{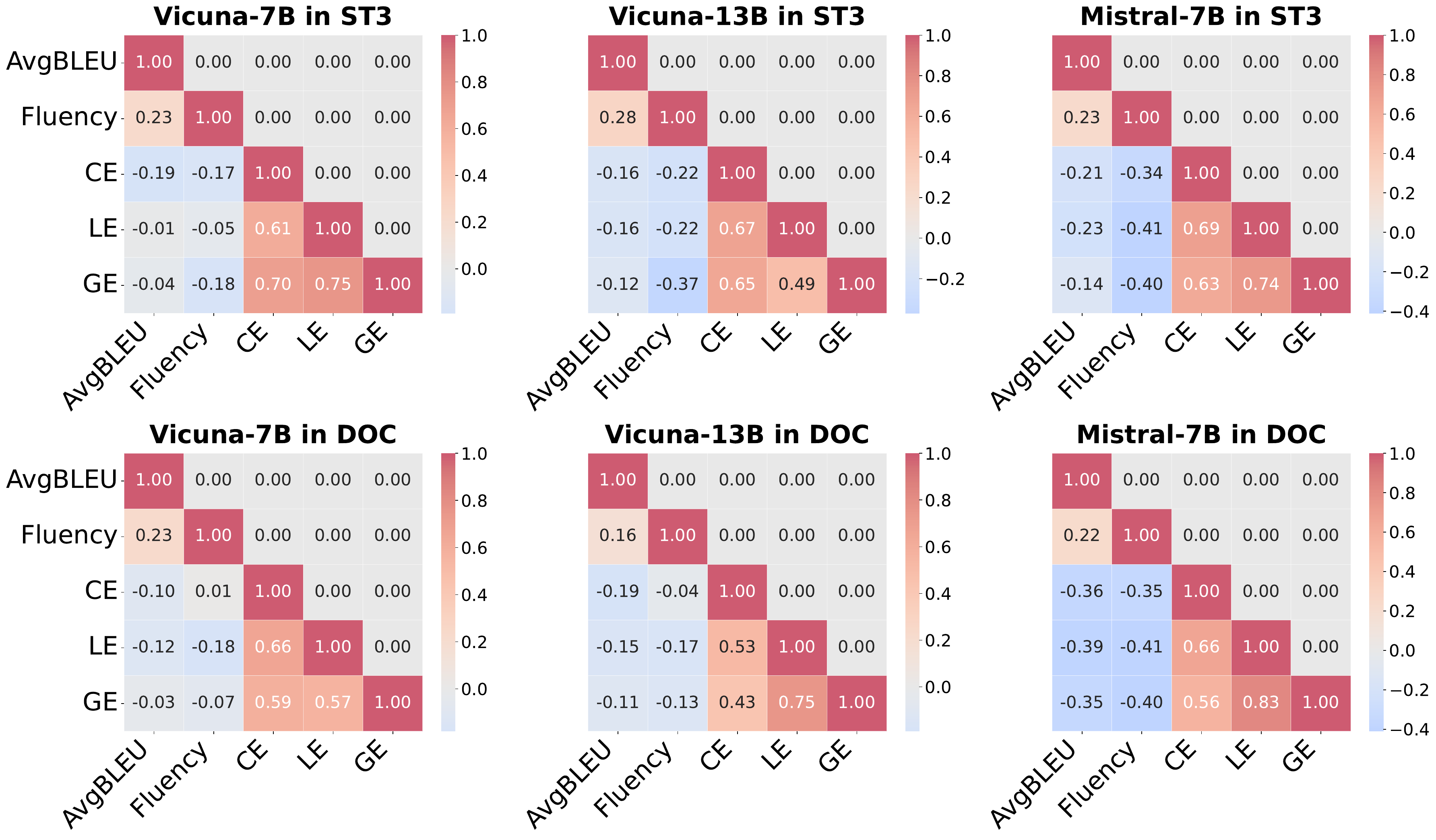}
    \caption{PCC Heatmaps among AvgBLEU, Fluency, CE(Content Errors), LE(Lexical Cohesion errors), GE(Grammatical Cohesion Errors) for Vicuna-7B, Vicuna-13B, and Mistral-7B under ST3 and DOC evaluation types in translation direction of \texttt{zh-en}.}
\end{figure*}

\begin{figure*}[h!]
    \centering
    \includegraphics[width=1\linewidth]{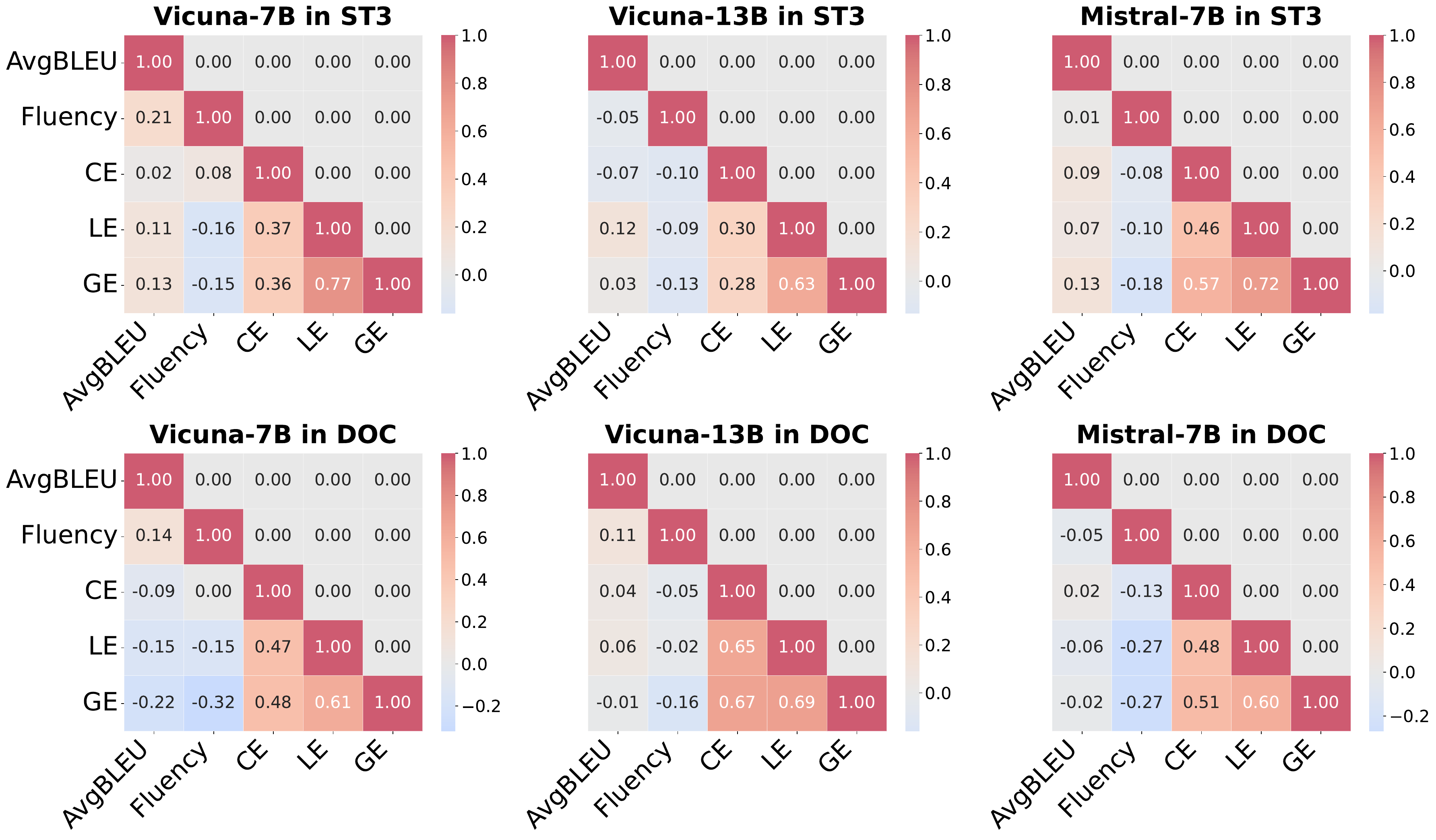}
    \caption{PCC Heatmaps among AvgBLEU, Fluency, CE(Content Errors), LE(Lexical Cohesion errors), GE(Grammatical Cohesion Errors) for Vicuna-7B, Vicuna-13B, and Mistral-7B under ST3 and DOC evaluation types in translation direction of \texttt{de-en}.}
\end{figure*}

\begin{figure*}[h!]
    \centering
    \includegraphics[width=1\linewidth]{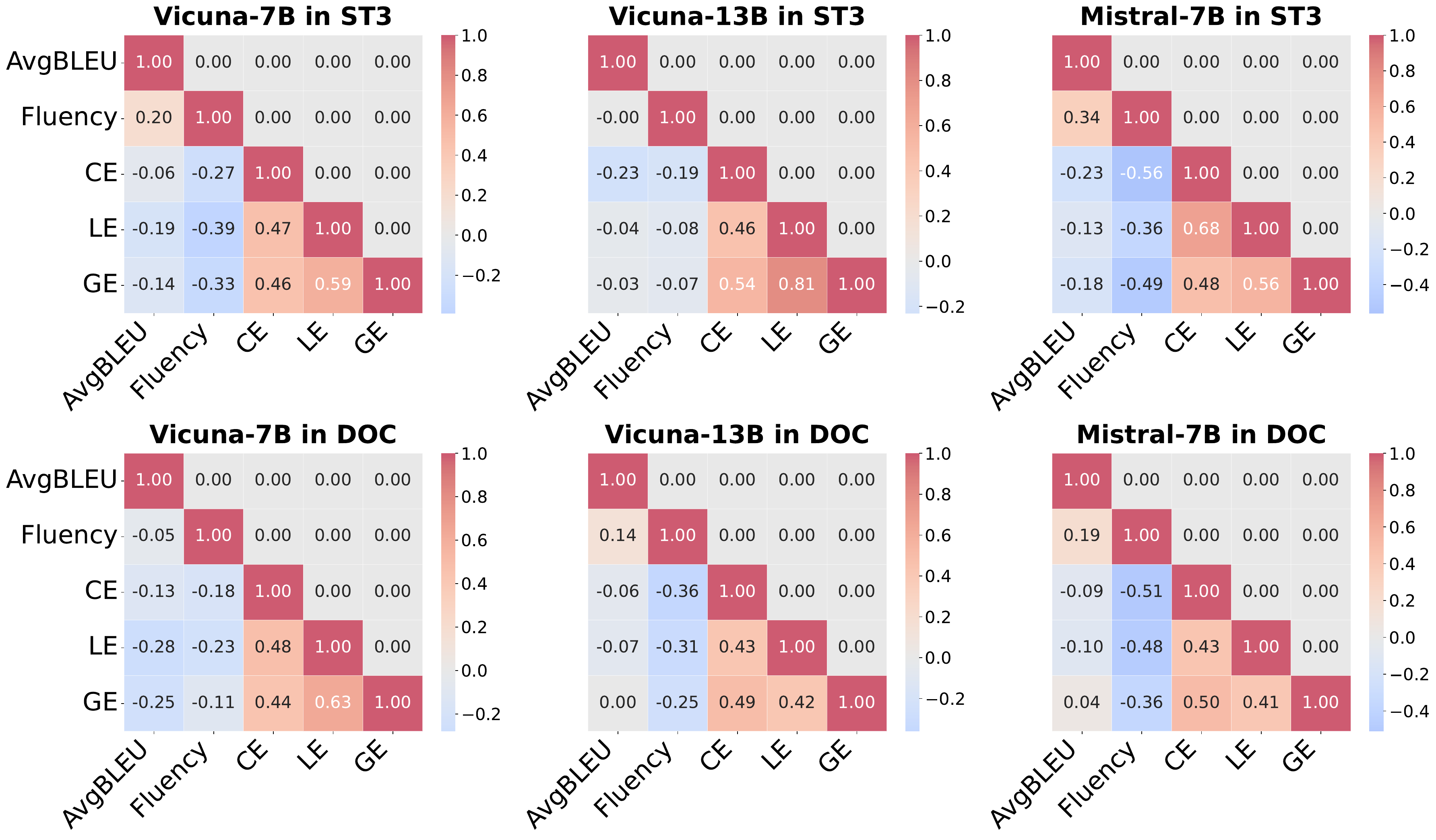}
    \caption{PCC Heatmaps among AvgBLEU, Fluency, CE(Content Errors), LE(Lexical Cohesion errors), GE(Grammatical Cohesion Errors) for Vicuna-7B, Vicuna-13B, and Mistral-7B under ST3 and DOC evaluation types in translation direction of \texttt{en-de}.}
\end{figure*}

\begin{figure*}[ht!]
    \centering
    \includegraphics[width=0.75\linewidth]{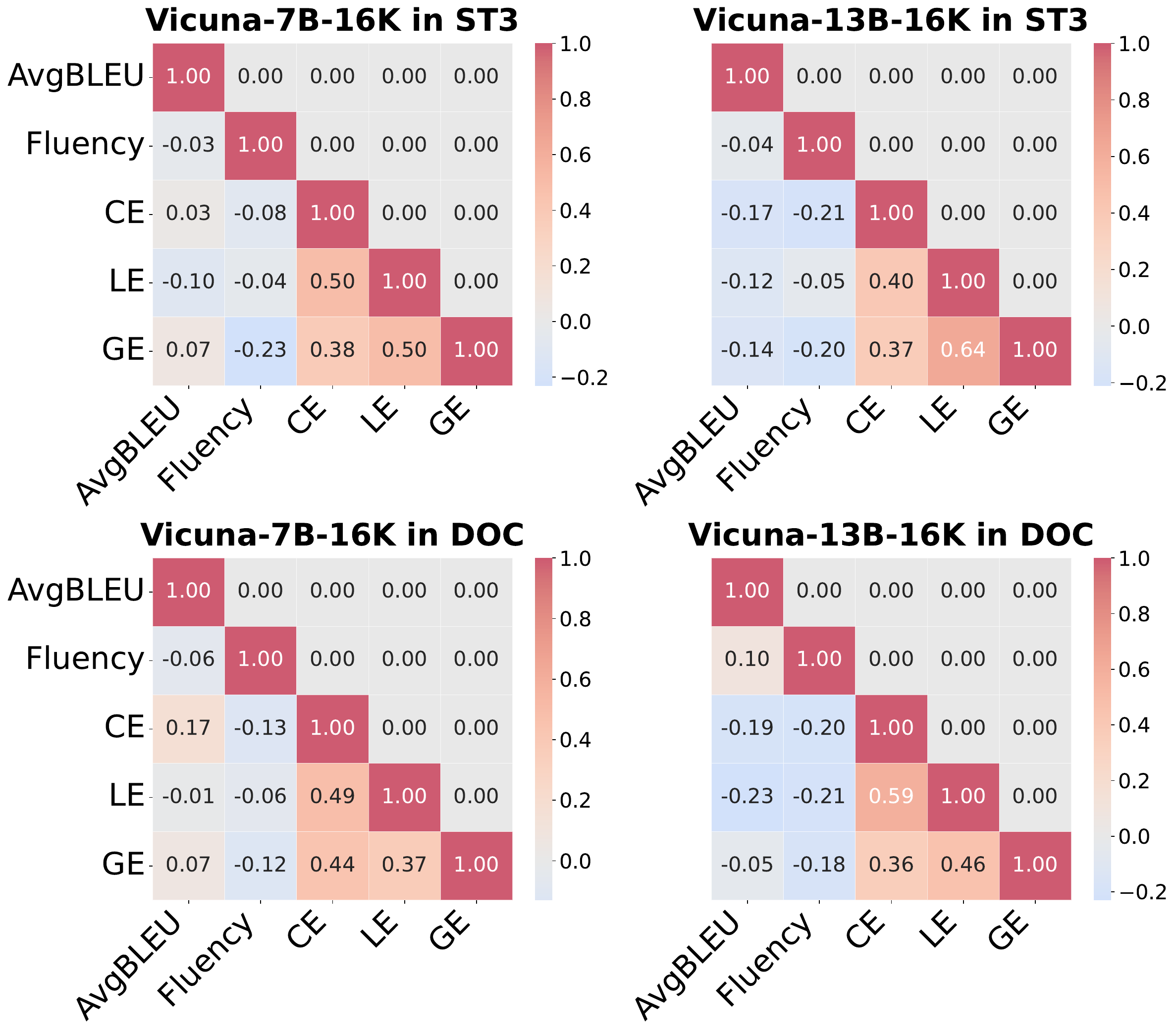}
    \caption{PCC Heatmaps among AvgBLEU, Fluency, CE, LE, GE for Vicuna-7B-16K and Vicuna-13B-16K under ST3 and DOC evaluation types in translation direction of \texttt{en-zh}.} 
\end{figure*}

\begin{figure*}[h!]
    \centering
    \includegraphics[width=0.75\linewidth]{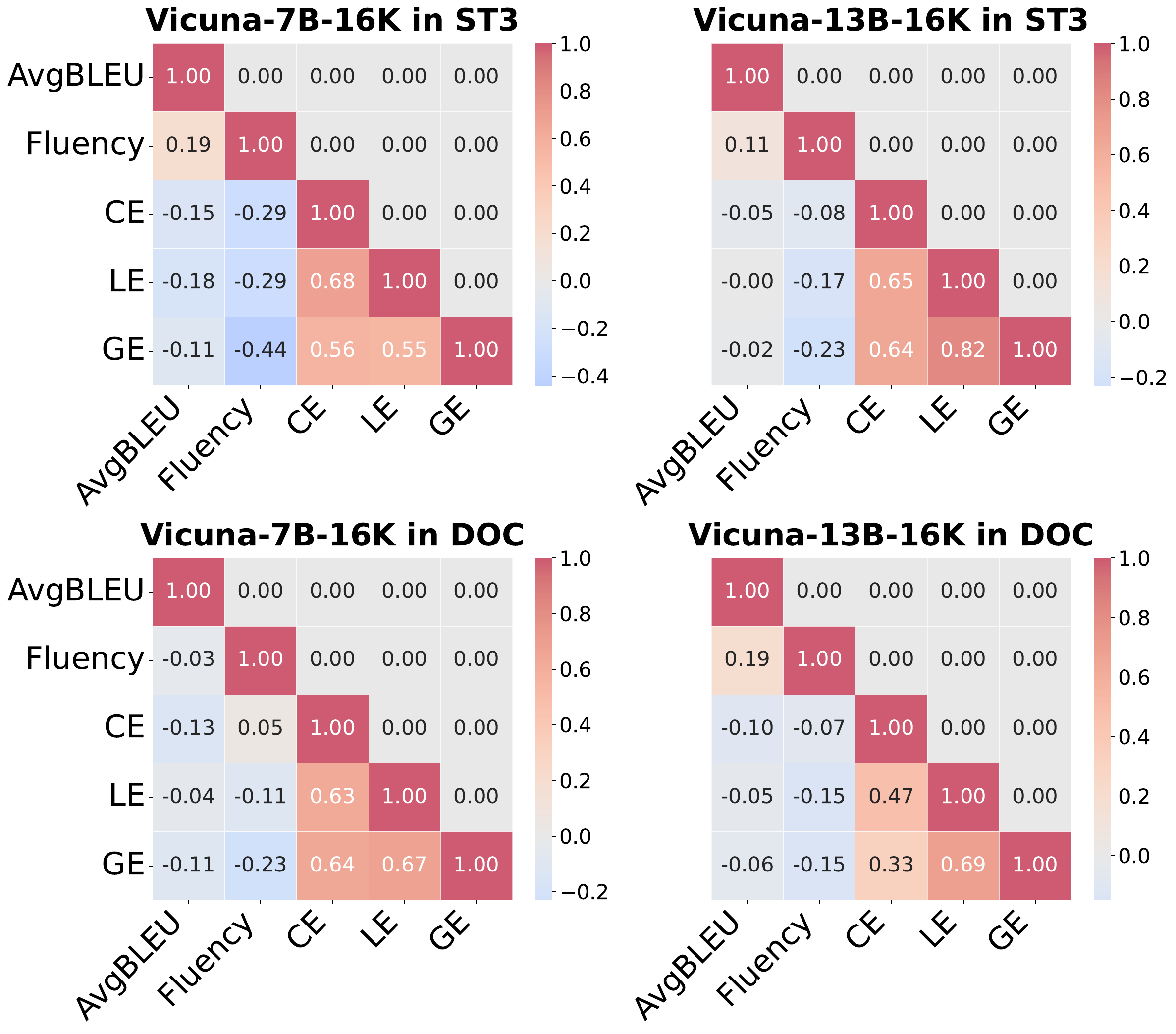}
    \caption{PCC Heatmaps among AvgBLEU, Fluency, CE(Content Errors), LE(Lexical Cohesion errors), GE(Grammatical Cohesion Errors) for Vicuna-7B-16K and Vicuna-13B-16K under ST3 and DOC evaluation types in translation direction of \texttt{zh-en}.}
\end{figure*}

\begin{figure*}[h!]
    \centering
    \includegraphics[width=0.75\linewidth]{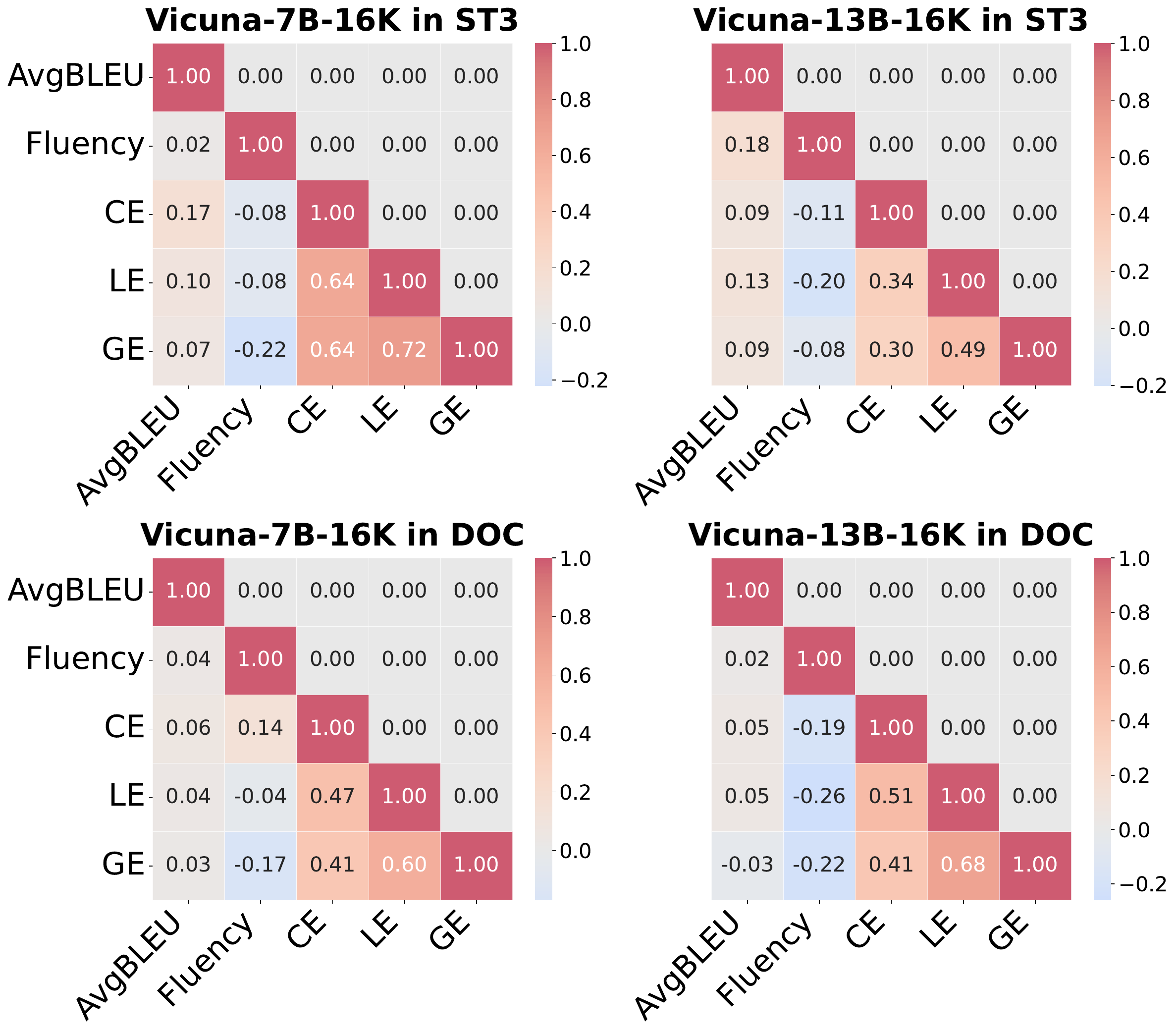}
    \caption{PCC Heatmaps among AvgBLEU, Fluency, CE(Content Errors), LE(Lexical Cohesion errors), GE(Grammatical Cohesion Errors) for Vicuna-7B-16K and Vicuna-13B-16K under ST3 and DOC evaluation types in translation direction of \texttt{de-en}.}
\end{figure*}

\begin{figure*}[h!]
    \centering
    \includegraphics[width=0.75\linewidth]{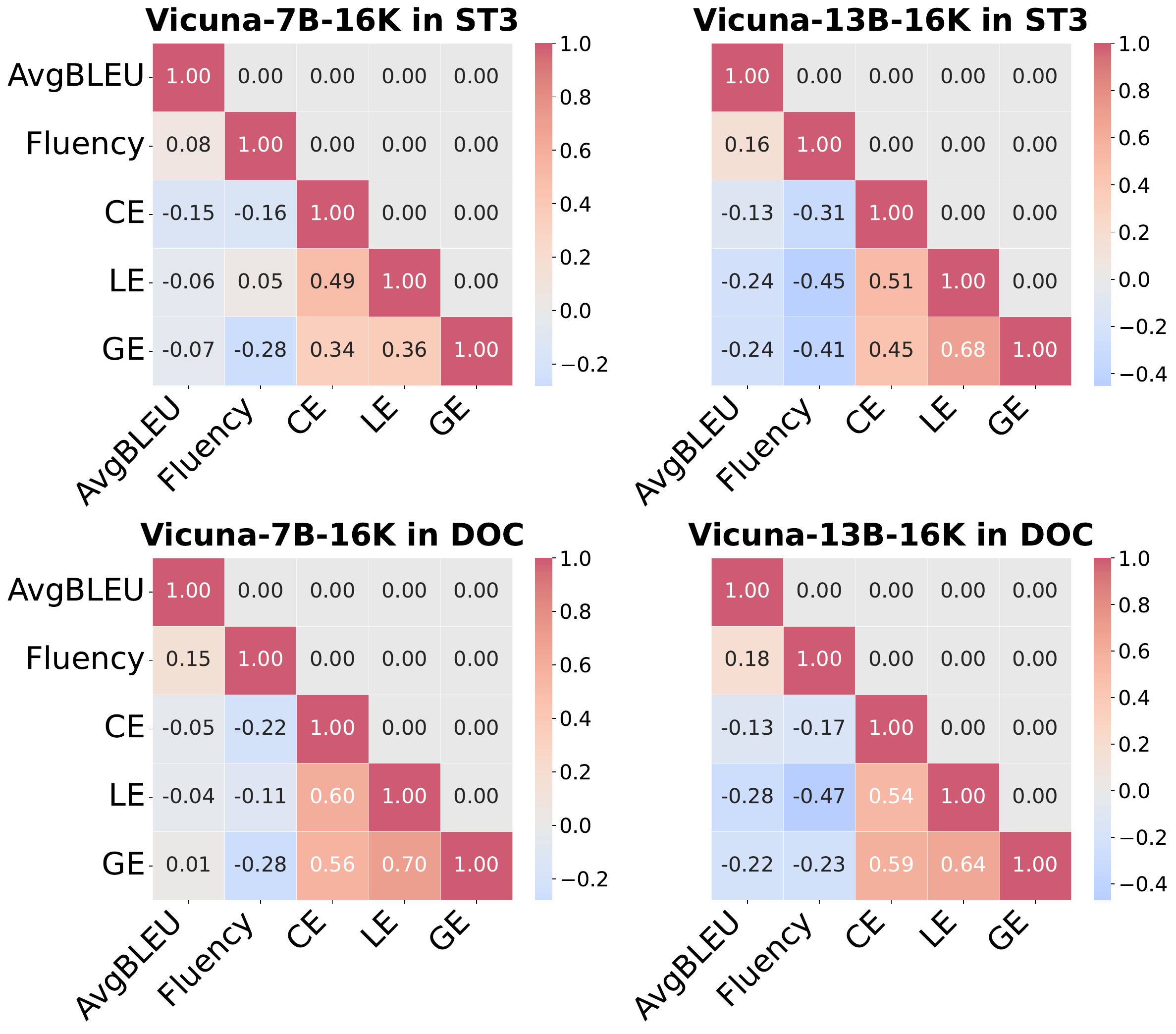}
    \caption{PCC Heatmaps among AvgBLEU, Fluency, CE(Content Errors), LE(Lexical Cohesion errors), GE(Grammatical Cohesion Errors) for Vicuna-7B-16K and Vicuna-13B-16K under ST3 and DOC evaluation types in translation direction of \texttt{en-de}.}
\end{figure*}

\end{document}